\documentclass[letterpaper, 10 pt, conference]{ieeeconf}  
\usepackage[T1]{fontenc}
\usepackage{xcolor}
\usepackage{tikz}
\usepackage{pgfplots}
\usepackage{multirow}
\usepackage{todonotes}
\usepackage{amsmath}
\usepackage{amssymb}
\usepackage{caption}

\usepackage{balance} 

\newcommand\x{\mathbf{x}}
\renewcommand\u{\mathbf{u}}
\renewcommand\o{\mathbf{o}}
\renewcommand\a{\mathbf{a}}
\newcommand\abs[1]{\left| #1 \right|}

\pgfplotsset{compat=newest}

\IEEEoverridecommandlockouts                              

\definecolor{deep1}{RGB}{76, 114, 176}
\definecolor{deep2}{RGB}{221, 132, 82}
\definecolor{deep3}{RGB}{85, 168, 104}
\definecolor{deep4}{RGB}{196, 78, 82}
\definecolor{deep5}{RGB}{129, 114, 179}
\definecolor{deep6}{RGB}{147, 120, 96}
\definecolor{deep7}{RGB}{218, 139, 195}
\definecolor{deep8}{RGB}{140, 140, 140}
\definecolor{deep9}{RGB}{204, 185, 116}
\definecolor{deep10}{RGB}{100, 181, 205}

\title{\LARGE \bf
Normalizing Flows are Capable Models for Bi-manual Visuomotor Policy
}

\author{
    Jialong Li$^{*1}$ \and
    Simon Kristoffersson Lind$^{*}$\and
    Wenrui Xie \and
    Maj Stenmark \and
    Volker Kr\"uger 
}

\begin{document}

\twocolumn[{%
\renewcommand\twocolumn[1][]{#1}%
\maketitle
\begin{center}
    \centering
    \includegraphics[width=0.99\textwidth]{figures/Pipeline.jpg}
    \captionof{figure}{Overview of the Normalizing Flow Policy. During training, the conditional normalizing flow learns a bijective mapping from the complex manifold of robot actions to a simple Gaussian prior $\mathbf{z}$, conditioned on visual observations $\mathbf{o}$.  At inference, the learned transformation is used inversely as $T^{-1}(\mathbf{z}|\mathbf{o})$ to generate actions.  A key advantage of normalizing flows is the ability to query the exact likelihood of generated actions, enabling two efficient inference optimisation strategies: 1) \emph{Stochastic Batch Selection}: generate a batch of candidates ($\mathbf{x}_{1 \dots N}$) and select the sample with the highest likelihood, and 2) \emph{Gradient Refinement}: refine a proposal trajectory by ascending the exact log-likelihood gradient $\nabla_{\mathbf{x}} \log p(\mathbf{x}|\mathbf{o})$ for a small number of steps.  The optimised trajectory is then executed by the robot.}
    \label{fig:teaser}
\end{center}%
 }]

\begingroup
\renewcommand\thefootnote{$^{*}$}
\footnotetext{Equal Contribution}
\endgroup

\begingroup
\renewcommand\thefootnote{$^{1}$}
\footnotetext{Corresponding author: \texttt{\footnotesize jialong.li@cs.lth.se}}
\endgroup

\begingroup
\renewcommand\thefootnote{}
\footnotetext{All authors with Department of Computer Science, Lund University LTH, Lund, Sweden, \textit{firstname.lastname@cs.lth.se}}
\endgroup

\begin{abstract}

The field of general-purpose robotics has recently embraced powerful probabilistic diffusion-based models to learn the complex embodiment behaviours. However, existing models often come with significant trade-offs, namely high computational costs for inference and a fundamental inability to quantify output uncertainty. 

We introduce \emph{Normalizing Flows Policy} (NF-P), a conditional normalizing flow–based visuomotor policy for bi-manual manipulation. NF-P learns a conditional density over action sequences and enables single-pass generative sampling with tractable likelihood computation. Using this property, we propose two inference-time optimization strategies: Stochastic Batch Selection, which selects the highest-likelihood trajectory among sampled candidates, and Gradient Refinement, which directly ascends the log-likelihood to improve action quality.

In both simulation and real robot experiments, NF-P achieves promising success rates compared to the baseline. In addition to improved task performance, NF-P demonstrates faster training and lower inference latency. These results establish normalizing flows as a competitive and computationally efficient visuomotor policy, particularly for real-time, uncertainty-aware robotic control.

\end{abstract}

\section{INTRODUCTION}

Learning from demonstrations has emerged as a promising paradigm for enabling robots to perform diverse manipulation tasks without extensive programming.  Recent approaches rely on powerful generative models such as diffusion models \cite{DiffusionPolicy,DDPM} to represent multi-modal distributions over action sequences conditioned on visual observations.  While diffusion models offer impressive expressiveness, they require expensive multi-step denoising during inference and do not provide tractable estimates of output likelihood.  Both limitations hamper their deployment on resource-constrained hardware and preclude a principled measure of confidence.

In this paper we argue that \emph{normalizing flows} (NFs) \cite{papamakarios} offer an attractive alternative to exisiting RGB-based visuomotor policy learning models.  NFs learn a bijective, differentiable transformation that maps complex action distributions to a simple latent prior and vice versa.  This bijection enables efficient sampling via a single forward pass and allows exact likelihood computation, enabling the policy to quantify uncertainty and optimise generated trajectories.  We present \emph{Normalizing Flow Policy} (NF-P), and demonstrate that it can match or exceed the performance of diffusion-based visuomotor policies while offering faster training, inference, and explicit confidence measures. NF-P brings together three advantages that, to the best of our knowledge, have not previously been demonstrated in previous robotic policy: (i) exact likelihood computation for principled uncertainty estimation, (ii) single-pass generative sampling that enables real-time inference, and (iii) the capacity to model complex distributions over high-dimensional bimanual action spaces.  This combination allows NF-P to provide qualitatively new capabilities such as uncertainty-guided action optimisation and safe exploration on real hardware.

Our contributions are threefold:
\begin{itemize}
    \item We propose NF-P, a conditional normalizing flow architecture tailored for RGB-based visuomotor policy learning, including two novel design choices: stride data sampling during training, and likelihood optimization over inference.
    \item We evaluate NF-P across a wide range of simulated bimanual tasks in the RoboTwin 2.0 framework \cite{RoboTwin2} as well as on real robot demonstrations. NF-P consistently achieves competitive or superior success rates compared to the baseline diffusion policy and exhibits strong data efficiency.
    \item We demonstrate how NF-P scales with the size of the dataset, demonstrating that NF-P has good scaling properties.
    \item We conduct an ablation study investigating factors that improve the performance of NF-P: stride-based sequencing, likelihood-based output optimisation, and sampling variance.
\end{itemize}
Taken together, we show that normalizing flows are highly capable for visuomotor policy learning,
while offering a substantial advantage over diffusion-based policies
in the form of efficient training and inference, and an explicit uncertainty estimate.

\section{RELATED WORK}

\subsection{Vision-based Imitation Learning}

Vision-based imitation learning aims to leverage the ubiquitous nature and high information density of visual data to map observations to robotic actions. Within this paradigm, Transformer-based regression architectures \cite{CARP,RT-1,VQ-Bet,Baku} represent a dominant methodology, utilizing self-attention mechanisms to perform direct or autoregressive action prediction. These models benefit from the Transformer's ability to capture long-range temporal dependencies across sequences of visual tokens, making them highly effective for structured, predictable tasks. However, these deterministic or quasi-deterministic frameworks often collapse when encountering multi-modal data distributions \cite{Implicit}. Because they typically minimize a mean squared error (MSE) loss, they tend to regress toward the mean of conflicting demonstrations, rather than capturing distinct, valid behavioral modes. Furthermore, they lack inherent probabilistic mechanisms for failure recovery or uncertainty quantification, which limits their adaptability in non-deterministic environments. This fundamental inability to represent complex, non-Gaussian action distributions necessitates a shift from point-estimation models toward more expressive generative frameworks.

\subsection{Diffusion-based Policy}

To resolve the limitations of traditional regression, the Diffusion Policy framework \cite{DiffusionPolicy} and broader diffusion-based generative models have been integrated into the control loop. These approaches shift the objective toward modeling a comprehensive conditional distribution over the action space, allowing the agent to represent multiple valid trajectories simultaneously. This framework has inspired numerous extensions.
For instance, Hierarchical Diffusion Policy \cite{HierarchicalDP} combines high-level diffusion planning with low-level control, while Causal Diffusion Policy \cite{CausalDP} incorporates causal transformers to refine temporal modeling. Despite their expressive power, these models face a significant trade-off in computational efficiency. While several works aim to reduce inference costs through one-step distillation \cite{one-stepDP}, real-time iterative diffusion \cite{real_time_DP}, or point-cloud-based optimizations \cite{3DDP}, they inherit the underlying challenge of unquantified uncertainty and lack exact likelihood estimations.

Notably, recent advancements have introduced flow-matching vision-language-action (VLA) models such as $\pi_0$ and $\pi_{0.5}$ \cite{PI0,pi0_5}. Although these models utilize flow-matching to achieve state-of-the-art performance in robotic manipulation, they are fundamentally designed as large-scale VLAs that rely on massive pre-trained foundation models. Given that this work prioritizes efficiency, such large-scale VLA architectures fall outside our primary scope.

\subsection{Normalizing Flows}
Normalizing flows have established a robust theoretical foundation in density estimation and generative modeling \cite{RealNVP} by leveraging the change-of-variables formula to map simple base distributions to complex target manifolds. Seminal architectures such as RealNVP \cite{RealNVP} and GLOW \cite{GLOW} utilize affine coupling layers to construct expressive, bijective transformations that maintain computational tractability. Unlike diffusion-based models, which require iterative stochastic refinement, normalizing flows facilitate exact likelihood evaluation and single-pass, deterministic sampling. 

Recent research has successfully demonstrated the versatility of normalizing flows across diverse domains, including high-fidelity image generation \cite{NFforGenerative}, behavioral cloning within reinforcement learning frameworks \cite{NF_for_RL}, and as specialized action-decoding heads for Vision-Language-Action models \cite{Nina}. Building upon prior research \cite{SimonSCIA, SimonIROS}, which utilized normalizing flows as an auxiliary mechanism for out-of-distribution (OOD) detection to enhance robustness in visual processing, we extend these insights to visuomotor policy learning and show that normalizing flows can serve as the primary end-to-end policy model.

\section{PRELIMINARIS}

\begin{figure*}[t]
    \centering
    \includegraphics[width=0.99\textwidth]{figures/Data\_Stride.jpg}
    \caption{For a training sample $i$ anchored at current sample $t=6$, we extract a history of observations (blue) and a future trajectory of actions (yellow) using a temporal stride $s$ (here $s=2$, with both the observation and prediction horizon as 4).  The sampling window slides by a single frame (shift $=1$) for the next sample ($t=7$).  This stride sampling approach ensures that while the internal structure of each sample is sparse (low frequency), the dataset coverage remains dense, allowing the model to initiate plans from any state.}
    \label{fig:stride}
\end{figure*}

\subsection{RGB-based Visuomotor Policy}

In its simplest form, RGB-based visuomotor policy learning can be expressed as a supervised learning problem consisting of observation–action pairs $(\o_t, \a_t)$, in which the goal is to predict the action $\a_t$ given the input image observation $\o_t$ at time~$t$.  Robotic tasks are often multi-modal with multiple, equally valid action sequences from a given state.  Accordingly, modern policy models aim to learn a distribution $p(\a_t | \o_t)$ rather than a deterministic mapping.

\subsection{Normalizing Flows}
Similar to diffusion models, normalizing flows learn to model a distribution $p(\x)$
by transforming data samples into Gaussian noise.
Instead of modeling a diffusion process, normalizing flows represent the entire transformation as a single function \cite{papamakarios}:
\begin{equation} \label{eq:T}
    \u = T(\x), \enskip \x \sim p(\x), \enskip \u \sim p_u(\u)
    \enskip .
\end{equation}
It should be noted that $p_u$ can be chosen freely,
but the most common choice is Gaussian noise $p_u = \mathcal{N}(\mathbf{0}, \mathbf{I})$.
Importantly, $T$ is a \textit{diffeomorphism},
which means that it is bijective with both $T$ and $T^{-1}$ being differentiable.
Naturally, the bijective property allows sampling by:
\[
    \x = T^{-1}(\u), \enskip \u \sim p_u(\u) \enskip .
\]
Additionally, both $T$ and $T^{-1}$ being differentiable enables direct computation of $p(\x)$:
\begin{equation} \label{eq:px}
    p(\x) = p_u(T(\x)) \abs{\det J_T(\x)} \enskip .
\end{equation}
Here, $J_T(\x)$ refers to the Jacobian matrix of $T(\x)$.

Constructing a $T$ that fulfills \eqref{eq:T} for any distribution $p(\x)$ is non-trivial,
and as such learning-based methods, usually neural networks, are the only viable option.
The primary challenge in normalizing flows is to design a neural network-based $T$
that is sufficiently expressive, while fulfilling the requirements for a
diffeomorphism in addition to having a tractable Jacobian determinant.

A common way to design $T$, pioneered by RealNVP \cite{RealNVP}, is \textit{coupling layers}.
Coupling layers first divide their input $\x = [x_1, x_2, \ldots, x_D]$ into two parts,
commonly:
\[
    \x_1 = [x_1, \ldots, x_{D/2}], \enskip
    \x_2 = [x_{D/2+1}, \ldots, x_D] \enskip .
\]
$\x_1$ is then transformed using an invertible and differentiable function $f_\theta(\x_1)$,
with parameters $\theta$ computed from $\x_2$.
Formally:
\begin{gather*}
    \x_1' = f_\theta(\x_1), \enskip
    \theta = \operatorname{NN}(\x_2) \enskip \tag{3} \label{eq:NN_line} \\
    \x' = \operatorname{concat}(\x_1', \x_2) \enskip .
    \label{Coupling layer}
\end{gather*}

In this formulation, $\operatorname{NN}$ has no restrictions,
and is usually a regular feed-forward neural network.
Various choices exist for the design of $f_\theta$,
arguably the most common being an element-wise linear transform:
\[
    f_\theta(\x_1) = \x_1 \cdot \alpha + \beta, \enskip
    \text{with} \enskip
    (\alpha, \beta) = \theta \enskip .
\]
In this work, however, we adopt Neural Spline Flows \cite{NeuralSplineFlows},
which constructs $f_\theta$ as a rational-quadratic spline.

\section{METHODOLOGY}\label{sec:Methodology}

Aiming to apply normalized flow to real-world dual-arm tasks, we propose several methods on top of the basic architecture to address the relatively higher noise levels in real-world data compared to simulation data, and the higher modalities in dual-arm tasks compared to single-arm tasks.

\subsection{Observation Encoding and Conditioning}
A pre-trained ResNet18 \cite{ResNet} is used to extract an embedding vector
from each image before it is passed to the normalizing flow.
This image embedding is supplemented with a vector representing the latest action,
and these together form our observation $\o_t$.
A conditional distribution can be modeled by adding a fixed conditioning vector to the $\operatorname{NN}$ function in the formulation \eqref{eq:NN_line} as $\theta = \operatorname{NN}(\x,  \o_t)$.

\subsection{Strided Sampling}

A persistent challenge in imitation learning is the modeling of unwanted motions, which often arise from the high-frequency noise inherent in human demonstrations. While adopting \textit{observation horizons} and \textit{action chunking} \cite{DiffusionPolicy} \cite{aloha} promotes temporal smoothness, these methods often fail to filter out the unconscious jitters and pauses characteristic of teleoperated data collection. Although post-processing techniques like low-pass filtering or downsampling can mitigate these artifacts, they risk discarding valuable behavioral patterns in real-world data. 

To increase focus on meaningful actions during training, without reducing training data density, we introduce a \textit{strided sampling} method. Given an observation history $\mathbf{o}_{t-m:t}$, we model a sliding window of $n$ future actions $\mathbf{a}_{t:t+n}$.  By applying a temporal stride $s$ to both input and output sequences, as illustrated in Fig.~\ref{fig:stride}, the model is forced to focus on more larger transitions between successive states, effectively bypassing high-frequency motions and minor stalling. Under this regime, the Normalizing Flow Policy learns the conditional distribution:
\[
    p(\a_t, \a_{t+s}, \ldots, \a_{t+n s} \;|\; \o_t, \o_{t-s}, \ldots, \o_{t-m s}).
\]
In our later experiments we show that this technique significantly improves performance in both simulated environments and real-world robotic experiments.
Additionally, different values for the stride $s$ are evaluated in our ablation study.
Note that, since actions during inference are produced by sampling from the NF,
the stride $s$ is not required, and thus we only apply strided sampling during training.

\subsection{Likelihood-based Output Optimization}
An advantage of normalizing flows is the ability to evaluate $p(\a|\o)$ exactly. We demonstrate that likelihood estimate can be further leveraged at inference time to refine action selection. 

To this end, we propose two optimization strategies that guide the model toward high-confidence trajectories without discarding the underlying stochasticity of the learned distribution. \\
\textbf{Strategy 1: Stochastic Batch Selection (SBS):} In this approach, we generate a parallel batch of $N=128$ candidate action sequences from the latent distribution. By evaluating the log-likelihood for each candidate, we select the sequence that yields the highest probability. This "best-of-$N$" sampling effectively filters out low-density outliers that may arise from the tails of the distribution, ensuring the robot executes the most representative mode of the demonstrated behavior. \\
\textbf{Strategy 2: Gradient Refinement (GR):} Alternatively, we perform a localized optimization of a sampled trajectory. Starting from an initial sample, we execute 10 iterations of gradient ascent on the log-likelihood $\log p(\mathbf{a} | \mathbf{o})$ using the Adam optimizer \cite{Adam}. This process "climbs" the manifold of the action space toward the nearest local maximum. Unlike diffusion models, which require a fixed number of denoising steps to reach a valid sample, this method treats the initial sample as a warm start, using the flow's gradient information to improve the policy's output. As shown in Fig. \ref{fig:teaser}, both methods can be used directly at run time.


\section{SIMULATION EXPERIMENTS }

\begin{figure}[t]
    \centering
    \includegraphics[width=0.9\columnwidth]{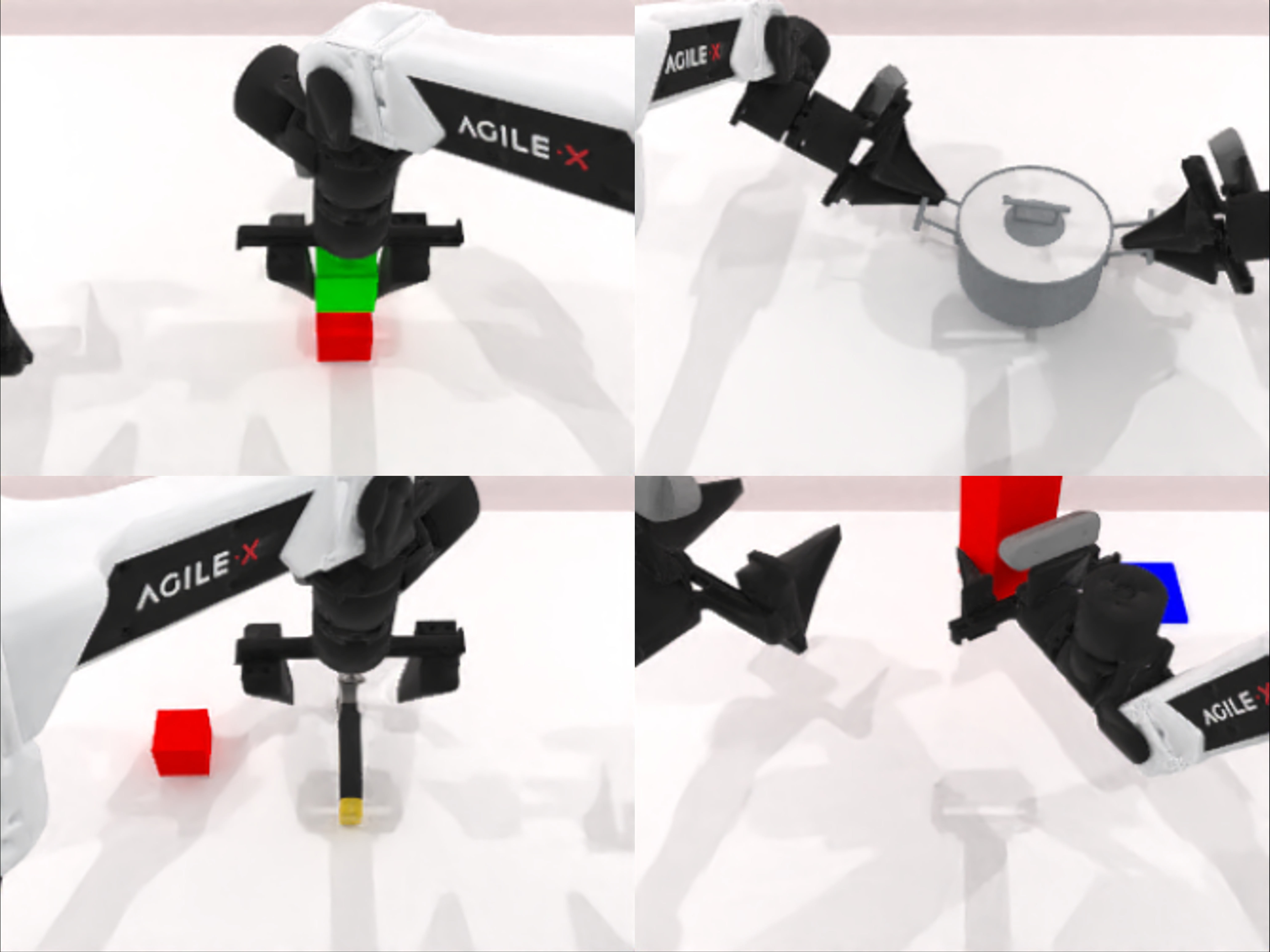}
    \caption{Visualization of four example simulation tasks: Stack Block Two, Lift Pot, Beat Block Hammer, and Handover Block.}
    \label{fig:sim_example}
\end{figure}

This section describes the evaluation protocol and results for NF-P on a diverse set of simulated bimanual tasks.  Unless otherwise noted, we report the \emph{success rate}, defined as the percentage of trials that achieve the task goal within a fixed horizon, and the \emph{inference time} per action sequence.  All reported metrics are averaged over 100 randomised test trials. Our evaluation covers 50 tasks from the RoboTwin~2.0 suite \cite{RoboTwin2}, which spans a wide range of bi-manual manipulation skills including grasping, lifting, tool use, handover, and stacking. Some example simulation tasks are shown in Fig. \ref{fig:sim_example}. 


\subsection{Experimental Setup}

We evaluate the performance of NF-P against the established Diffusion Policy (DP) baseline provided by RoboTwin-2.0, with scores taken from the official RoboTwin 2.0 leaderboard \cite{RTLeaderboard}.
For each task, NF-P is trained on 50 clean scripted demonstrations, using only the head camera as visual input. 
Our NF consists of 10 sequential coupling spline layers,
which results in approximately 100 million total trainable parameters.
Each NF is trained for 300 epochs with a fixed learning rate$=10^{-4}$ using the Adam optimizer\cite{Adam}.
We use a default value of $s=4$ for the strided sampling, justified by our ablation study.




\subsection{Comprehensive RoboTwin Evaluation}

\begin{figure*}[t]
    \centering
    \includegraphics[width=1.0\textwidth]{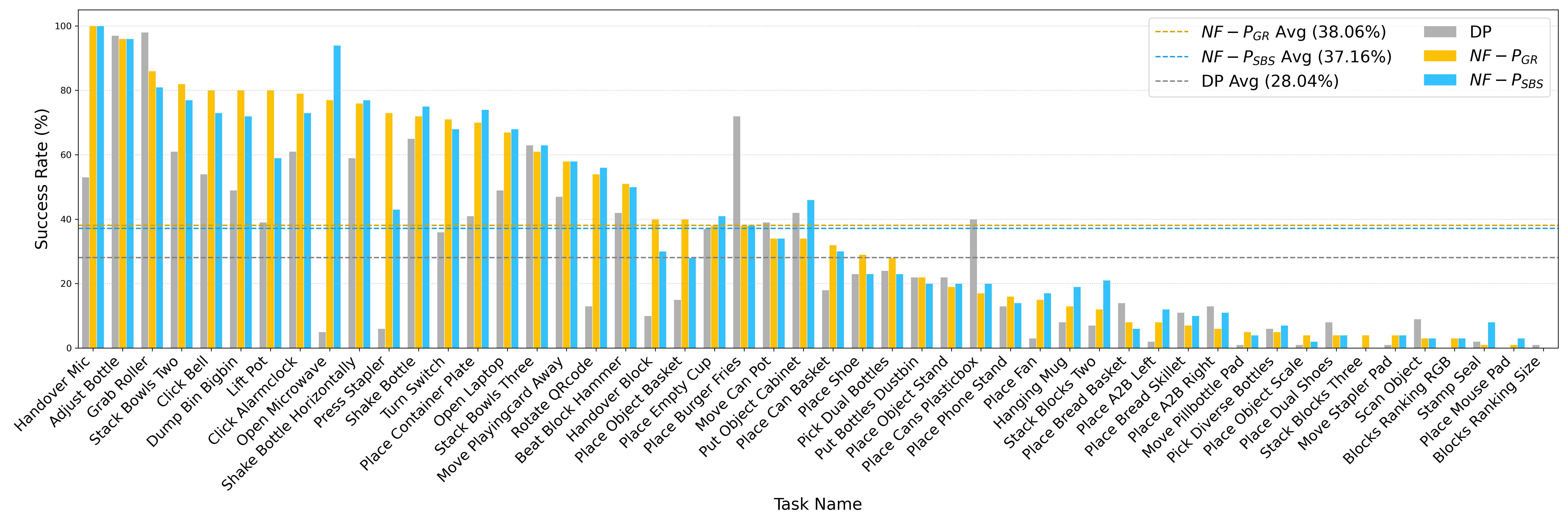}
    \caption{Summary of success rates on the full RoboTwin~2.0 evaluation suite. Bars show the success rate for 100 trials across all tasks for each method (DP, NF-P$_{\text{GR}}$ and NF-P$_{\text{SBS}}$). Normalizing flow policies consistently outperform the diffusion-policy baseline.}
    \label{fig:sim_result}
\end{figure*}

Evaluation results are summarised in Figure~\ref{fig:sim_result}. The comprehensive evaluation across the RoboTwin-2.0 benchmark suite reveals that NF-P outperforms the DP baseline in terms of overall success rates. As outlined in Fig.\ref{fig:sim_result}, NF-P with gradient refinement ($\text{NF-P}_{\text{GR}}$) achieves an average success rate of 38.06\%, representing a substantial 10 percentage-point improvement over the DP baseline (28.04\%). Notably, the performance gains are most pronounced in tasks requiring high precision and contact-rich interactions, such as Open Microwave (77\% vs. 5\%) and Handover Mic (100\% vs. 53\%). These results suggest that the exact density estimation provided by NFs allows the policy to better exploit the underlying demonstration manifold compared to DP.

Furthermore, the comparison between our two proposed inference strategies highlights the utility of likelihood-based refinement. While $\text{NF-P}_{\text{SBS}}$ provides a strong stochastic baseline, the GR variant frequently achieves better success rates in tasks like Press Stapler (73\% vs. 43\%) and Lift Pot (80\% vs. 59\%). This indicates that gradient ascent in log-likelihood is an effective mechanism for suppressing low-probability, suboptimal action sequences. Interestingly, in highly complex tasks like Hanging Mug or Open Microwave, $\text{NF-P}_{\text{SBS}}$ occasionally outperforms $\text{NF-P}_{\text{GR}}$, suggesting that a broad stochastic search may sometimes be more robust. Overall, both methods show promising results.

\section{REAL ROBOT EXPERIMENTS AND RESULTS}
This section details the physical hardware configuration, the selected manipulation tasks, and the subsequent performance results of NF-P in a bi-manual real-world environment

\subsection{Experiment Setup}
We evaluate NF-P on a bi-manual Kuka setup, which contains two Kuka iiwa 7 robots. Similar to the simulated experiments, only a single Realsense D435I is used, which captures the whole scene, as highlighted in Fig. \ref{fig:kukas}. Task data are collected using Quest2ROS2 \cite{quest2ros2}, a VR based teleoperation system, and the data types are the same as the RoboTwin 2.0 simulations detailed above.

\subsection{Task Definitions and Policy Training}
To assess the robustness and coordination of NF-P, we selected two distinct tasks that challenge different aspects of robotic control. \\
\textbf{Stack Blocks Two}: This task requires the robot to sequentially pick up two colored blocks and stack them in a predefined order. Beyond simple reaching and grasping, this task evaluates the model's long-horizon temporal reasoning and the precision required to ensure the resulting stack remains physically stable. \\
\textbf{Towel Folding}: Unlike rigid-body manipulation, this task involves a deformable object that requires precise bi-manual coordination. The two robotic arms must execute a synchronized grasp, move at a uniform velocity to maintain tension, and perform a simultaneous release.

For the training phase, we collected 150 demonstrations for Stack Block Two and 100 for Towel Folding. To compensate for the higher noise levels in real robot data, the NF-P architecture was configured with an observation and prediction horizon of 4 and a temporal stride $s=6$, and uses SBS optimization in inference. Similarly, the DP baseline employed the standard DP-C implementation \cite{DiffusionPolicy}, which has a longer horizon than the simulation setup. 

\subsection{Results and Discussion}

\begin{figure*}[t]
    \centering
    \includegraphics[width=1.0\textwidth]{figures/Kuka.jpg} 
    \caption{Real-world experimental setup and task progression. (Left) The dual-arm manipulation environment for the Stack Block Two (top row) and Towel Folding (bottom row) tasks. The wrist and head cameras are not used for both task, the red circle indicates the single scene camera which provides visual observation input for the models.
    (Right) Key terminal stages of the tasks viewed from the perspective of the scene camera. These stages correspond to the bottleneck evaluation milestones (e.g., picking, placing, and folding).}
    \label{fig:kukas}
\end{figure*}

\begin{table*}[ht]
\centering
\begin{tabular}{ccccccc}
\hline
                & \multicolumn{6}{c}{Stack Block Two}                                                                                                                                                                              \\ \hline
Method          & \multicolumn{1}{c|}{Initial Fail} & \multicolumn{1}{c|}{Arm One Picking} & \multicolumn{1}{c|}{Arm One Placing} & \multicolumn{1}{c|}{Arm Two Picking} & \multicolumn{1}{c|}{Arm Two Placing} & Success Rate(\%) \\
DP              & \multicolumn{1}{c|}{7}            & \multicolumn{1}{c|}{0}              & \multicolumn{1}{c|}{10}               & \multicolumn{1}{c|}{0}               & \multicolumn{1}{c|}{3}               & 15               \\
NF-P$_{\text{multi}}$ & \multicolumn{1}{c|}{0}            & \multicolumn{1}{c|}{0}               & \multicolumn{1}{c|}{1}               & \multicolumn{1}{c|}{7}               & \multicolumn{1}{c|}{12}              & \textbf{60}               \\ \hline
                & \multicolumn{6}{c}{Towel Folding}                                                                                                                                                                                \\ \hline
Method          & \multicolumn{1}{c|}{Initial Fail} & \multicolumn{2}{c|}{Towel Picking}                                          & \multicolumn{2}{c|}{Towel Folding}                                 & Success Rate(\%) \\
DP              & \multicolumn{1}{c|}{9}            & \multicolumn{2}{c|}{0}                                                      & \multicolumn{2}{c|}{11}                                                     & 55               \\
NF-P$_{\text{multi}}$ & \multicolumn{1}{c|}{3}            & \multicolumn{2}{c|}{2}                                                      & \multicolumn{2}{c|}{15}                                                     & \textbf{75}              
\end{tabular}
\caption{Bottleneck Analysis and Success Rates for Real-World Rollouts. The table records the terminal stage of each of 20 trials alongside the overall success rate. While DP frequently succumbs to early-stage bottlenecks (e.g., failing at initial grasp or prior to execution), our method reliably completes early sub-tasks, pushing failures to later stages and achieving higher overall success rates (60\% and 75\%, respectively).}
\label{table:real_rollout}
\end{table*}

To highlight the performance differences observed during real-world execution, we conducted a rigorous bottleneck analysis focused on the terminal stages of each multi-stage task. As detailed in Table \ref{table:real_rollout}, this analysis identifies specific points of failure, revealing how each policy handles sequential dependencies and compounding errors across long-horizon trajectories.

A critical behavioral trend emerges when examining the DP baseline: it frequently fails to initialize execution, and it often gets stuck between sub-goals. In the Stack Block Two task, DP suffered from an initial failure rate of 35\%, often failing to trigger the first picking motion. Furthermore, 50\% of its trials terminated prematurely after the "Arm One Placing" stage. Our observations indicate that DP commonly fails when transitioning between sub-goals. For example, after the first block has been placed,
instead of engaging the second arm to place the other block,
DP would often get stuck in a cycle repeating the first arm's trajectory.
Similarly, in the Towel Folding task, DP exhibited initiation bottlenecks, with 45\% of trials failing at the onset.
In these instances, 
the model is typically able to steer the robot to target object but fails to give the gripper-close command,
stalling indefinitely. 



While NF-P also exhibits some amount of stalling when transitioning between sub-goals,
the average transition time is significantly lower.
Throughout our experiments, DP was given an 8-minute window for each sub-goal transition,
whereas NF-P achieved a higher success rate with only a 30-second limit.

In the Stack Block Two task, NF-P successfully picked and placed the first block in 95\% of trials, and achieved a 0\% initial failure rate. The sparse failures encountered were significantly deeper into the execution horizon compared to DP, primarily occurring at the "Arm Two Picking" stage, due to missing the placement of the block at the last step. For Towel Folding, NF-P initiated the trajectory successfully in 85\% of cases, and only failed during towel picking in 10\% of the cases.

These results suggest that the exact likelihood estimation inherent in Normalizing Flows provides a more stable density manifold for the agent to follow. By effectively mitigating early-stage stochastic drift and facilitating rapid state transitions, NF-P ensures that the policy remains within distribution even during complex handovers between sub-goals. This robustness leads to markedly higher overall success rates compared to DP, validating the efficacy of normalizing flows based architectures for extended bi-manual robotic manipulations.

\section{DATA SCALING EXPERIMENT}
One concern regarding NFs, is that they might not be as expressive as other models,
as they are constrained by requiring a tractable inverse and Jacobian determinan \cite{papamakarios}.
In order to verify that NFs are capable of scaling to larger datasets, we conduct an additional experiment using the RoboTwin 2.0 simulation framework.
Here, we select 4 different tasks: Beat Block Hammer, Lift Pot, Handover Block, and Stack Blocks Two.
For each task, we train and evaluate both NF-P and DP (configuration DP-C \cite{DiffusionPolicy}) using a varying number of training episodes from 10 up to 500.
Our results, in Fig.~\ref{fig:episodes_experiment} show that NF-P exhibit similar scaling properties as DP across all 4 tasks, alleviating any initial concerns.

\begin{figure*}[t]
    \begin{tabular}{cccc}
    \multicolumn{4}{c}{
    \begin{tikzpicture}
    \node[draw] {
        \begin{tabular}{ccc}
            NF-P$_\text{SBS}$: \ref{nf_multi_bbh} &
            NF-P$_\text{GR}$: \ref{nf_bbh} &
            DP: \ref{dp_bbh}
        \end{tabular}
    };
    \end{tikzpicture}
    }
    \\
    \begin{tikzpicture}
    \begin{axis}[
        width=0.22\paperwidth,  
        height=0.13\paperheight,
        title={\scriptsize Beat Block Hammer},
        xmode=log,
        ylabel={Success \%},
        y label style={font=\footnotesize},
        ytick={0, 50, 100},
        y tick label style={font=\scriptsize},
        ytick pos=left,
        ymin=-10,
        ymax=110,
        xlabel={\#train episodes},
        x label style={font=\footnotesize},
        xtick={10,25,50,100,200,300,500},
        x tick label style={font=\scriptsize, rotate=90},
        xticklabels={10, 25, 50, 100, 200, 300, 500},
        xtick pos=bottom,
        every axis plot/.append style={
            mark=*, line width=0.2pt, mark size=1pt
        },
    ]
    \addplot [color=deep1] coordinates {
        (10,13) (25,40) (50,51) (100,67) (200,83) (300,85) (500,96)
    }; \label{nf_bbh}
    \addplot [color=deep3] coordinates {
        (10,13) (25,39) (50,50) (100,68) (200,75) (300,85) (500,91)
    }; \label{nf_multi_bbh}
    \addplot [color=deep2] coordinates {
        (10,12) (25,16) (50,48) (100,81) (200,62) (300,96) (500,99)
    }; \label{dp_bbh}
    \end{axis}
    \end{tikzpicture}
    &
    \begin{tikzpicture}
    \begin{axis}[
        width=0.22\paperwidth,
        height=0.13\paperheight,
        title={\scriptsize Lift Pot},
        xmode=log,
        y label style={font=\footnotesize},
        ytick={0, 50, 100},
        y tick label style={font=\scriptsize},
        ytick pos=left,
        ymin=-10,
        ymax=110,
        xlabel={\#train episodes},
        x label style={font=\footnotesize},
        xtick={10,25,50,100,200,300,500},
        x tick label style={font=\scriptsize, rotate=90},
        xticklabels={10, 25, 50, 100, 200, 300, 500},
        xtick pos=bottom,
        every axis plot/.append style={
            mark=*, line width=0.2pt, mark size=1pt
        },
    ]
    \addplot [color=deep1] coordinates {
        (10,18) (25,39) (50,80) (100,95) (200,98) (300,100) (500,100)
    }; \label{nf_lp}
    \addplot [color=deep3] coordinates {
        (10,2) (25,35) (50,59) (100,94) (200,100) (300,100) (500,100)
    }; \label{nf_multi_lp}
    \addplot [color=deep2] coordinates {
        (10,14) (25,26) (50,37) (100,82) (200,100) (300,100) (500,100)
    }; \label{dp_lp}
    \end{axis}
    \end{tikzpicture}
    &
    \begin{tikzpicture}
    \begin{axis}[
        width=0.22\paperwidth,
        height=0.13\paperheight,
        title={\scriptsize Handover Block},
        xmode=log,
        y label style={font=\footnotesize},
        ytick={0, 50, 100},
        y tick label style={font=\scriptsize},
        ytick pos=left,
        ymin=-10,
        ymax=110,
        xlabel={\#train episodes},
        x label style={font=\footnotesize},
        xtick={10,25,50,100,200,300,500},
        x tick label style={font=\scriptsize, rotate=90},
        xticklabels={10, 25, 50, 100, 200, 300, 500},
        xtick pos=bottom,
        every axis plot/.append style={
            mark=*, line width=0.2pt, mark size=1pt
        },
    ]
    \addplot [color=deep1] coordinates {
        (10, 10) (25, 25) (50, 40) (100, 63) (200, 82) (300, 72) (500, 81)
    }; \label{nf_hb}
    \addplot [color=deep3] coordinates {
        (10, 0) (25, 24) (50, 30) (100, 68) (200, 73) (300, 78) (500, 72)
    }; \label{nf_multi_hb}
    \addplot [color=deep2] coordinates {
        (10, 1) (25, 20) (50, 44) (100, 13) (200, 77) (300, 89) (500, 95)
    }; \label{dp_hb}
    \end{axis}
    \end{tikzpicture}
    &
    \begin{tikzpicture}
    \begin{axis}[
        width=0.22\paperwidth,
        height=0.13\paperheight,
        title={\scriptsize Stack Block Two},
        xmode=log,
        y label style={font=\footnotesize},
        ylabel style={yshift=4pt},
        ytick={0, 30, 60},
        y tick label style={font=\scriptsize},
        ytick pos=left,
        ymin=-10,
        ymax=70,
        xlabel={\#train episodes},
        x label style={font=\footnotesize},
        xtick={10,25,50,100,200,300,500},
        x tick label style={font=\scriptsize, rotate=90},
        xticklabels={10, 25, 50, 100, 200, 300, 500},
        xtick pos=bottom,
        every axis plot/.append style={
            mark=*, line width=0.2pt, mark size=1pt
        },
    ]
    \addplot [color=deep1] coordinates {
        (10, 0) (25, 8) (50, 12) (100, 38) (200, 55) (300, 59) (500, 66)
    }; \label{nf_sbt}
    \addplot [color=deep3] coordinates {
        (10, 0) (25, 6) (50, 21) (100, 39) (200, 56) (300, 55) (500, 64)
    }; \label{nf_multi_sbt}
    \addplot [color=deep2] coordinates {
        (10, 0) (25, 5) (50, 6) (100, 21) (200, 35) (300, 59) (500, 53)
    }; \label{dp_sbt}
    \end{axis}
    \end{tikzpicture}
    \end{tabular}
    \caption{Results for scaling simulated experiments with varying amounts of training data.}
    \label{fig:episodes_experiment}
\end{figure*}

\section{MODEL EFFICIENCY}


Computational overhead during both training and inference is a critical constraint for real-world robotic deployment. NF-P demonstrates significant advantages in this regard, achieving high-fidelity results with relatively minimal resource expenditure. Specifically, on a single NVIDIA RTX 4090, the model reaches convergence in as few as 300 epochs on a dataset of 100 demonstrations, which takes roughly 50 mins to train. Also, due to its single-pass, deterministic sampling architecture, $\text{NF-P}_{\text{SBS}}$ achieves inference times under 20ms. This represents a substantial improvement over diffusion models, enabling high-frequency control loops that are essential for reactive manipulation. Even the GR variant ($\text{NF-P}_{\text{GR}}$), which performs 10 steps of gradient ascent, maintains a latency of 455ms, which is faster than the standard DP baseline. Notably, $\text{NF-P}_{\text{SBS}}$ offers an ideal middle ground, retaining the real-time advantages of the flow backbone while delivering the performance gains associated with likelihood-based selection. This efficiency ensures that NF-P can be deployed on standard robotic hardware without the need for specialized high-latency compensators.

\section{Ablation Study}
To isolate the contributions of specific design choices within the NF-P framework, we conducted a systematic ablation study using the Beat Block Hammer simulation task. We evaluated the sensitivity of the model to the temporal stride $s$, the sampling variance $\sigma$, and the choice of output optimization strategy. The results, summarized in Table \ref{table:ablation}, identify the configuration of $s=4$, $\sigma=0.5$ with gradient-ascent optimization as the optimal parameter set, achieving a peak success rate of 51\% on the RoboTwin 2.0 Beat Block Hammer task.

Both $\text{NF-P}_{\text{SBS}}$ and $\text{NF-P}_{\text{GR}}$ yield substantial performance gains over direct sampling $\text{NF-P}$. This underscores the benefits of guiding the stochastic output toward high-probability regions of the learned manifold to ensure precision.

Temporal stride appears to play a critical role in balancing the generated motions. A stride of $s=1$ resulted in a lower success rate of 37\%. Conversely, excessively large strides ($s=8$) also degraded performance to 46\%. The value $s=4$ appears to be the optimal for filtering demonstration noise while maintaining behavioral fidelity.

A moderate sampling variance appears to aid task execution, with $\sigma=0.5$ delivering the highest success rate.
Standard sampling, with $\sigma=1$ yields reduced performance.
Similarly, decreasing it to $\sigma=0.25$ also led to worse success rates, indicating that the policy requires a moderate level of stochastic exploration during the inference process to remain robust to environmental perturbations.

Predicting a single action instead of an action chunk reduced the success rate to 28\%. This result reinforces the importance of action chunking with regard to temporal consistency.

\begin{table}[t]
    \centering
    \caption{NF-P ablations on the RoboTwin 2.0 \emph{Beat Block Hammer} task.  Each entry reports the success rate (\%) from 100 trials when varying one component of the NF-P design.  The default configuration is $s=4$, $\sigma=0.5$, and gradient refinement.}
    \label{table:ablation}
    \begin{tabular}{ccc|c}
    \multicolumn{3}{c|}{\textbf{Ablation}} & \textbf{Success rate} \\
    \hline
    NF-P$_{\text{GR}}$ & $s=4$ & $\sigma=0.5$ & \textbf{51} \\
    NF-P$_{\text{SBS}}$ & $s=4$ & $\sigma=0.5$ & 50 \\
    NF-P & $s=4$ & $\sigma=0.5$ & 37 \\
    NF-P$_{\text{GR}}$ & $s=8$ & $\sigma=0.5$ & 46 \\
    NF-P$_{\text{GR}}$ & $s=2$ & $\sigma=0.5$ & 48 \\
    NF-P$_{\text{GR}}$ & $s=1$ & $\sigma=0.5$ & 37 \\
    NF-P$_{\text{GR}}$ & $s=4$ & $\sigma=1.0$ & 41 \\
    NF-P$_{\text{GR}}$ & $s=4$ & $\sigma=0.75$ & 47 \\
    NF-P$_{\text{GR}}$ & $s=4$ & $\sigma=0.25$ & 42 \\
    NF-P & \multicolumn{2}{c|}{No action chunking} & 28
    \end{tabular}
\end{table}


\section{CONCLUSION}
In this work, we introduced Normalizing Flow Policy (NF-P), a visuomotor policy framework based on normalizing flows designed for complex, bi-manual manipulation. By utilizing normalizing flows, NF-P provides a unique combination of exact likelihood estimation and efficient single-pass inference.

Our comprehensive evaluation across the RoboTwin-2.0 benchmark suite demonstrates that NF-P consistently matches or exceeds the performance of Diffusion Policy (DP), achieving a higher average success rate while requiring significantly fewer training epochs. Real-world experiments on a bi-manual Kuka setup further validate the model's robustness, specifically its ability to overcome stalling and transition bottlenecks.

Through detailed ablation studies, we identified that stride-based sampling and likelihood-based optimization lead to improved perfomance. Ultimately, NF-P represents a mathematically sound and computationally efficient alternative for scaling robotic learning to multi-modal settings where real-time responsiveness and reliable uncertainty quantification are required.

\section{FUTURE WORK}
While we have explored several different techniques for making  NFs viable and even competitive in the domain of visuomotor policy learning, our work is by no means exhaustive.
One such method that we left unexplored is the addition of noise during training \cite{NFforGenerative}.
It would be an interesting direction to investigate how adding noise behaves in conjunction with out gradient ascent likelihood-optimization, since the noise removal is effectively equivalent to one step of gradient ascent.

There are also works, for example \cite{real_time_DP}, that attempt to improve inference times in diffusion models by updating an existing action sequence using fewer denoising steps compared to generating actions from scratch.
Similar approaches could be explored for NFs. However, in the case of NFs, we believe such approaches would rather be applicable for improving consistency and smoothness in the action sequences, since the NF's single-pass inference already provides low latency.

Another potential research direction could be to evaluate different NFs architectures. In this work, we adopted coupling layers due to their fast inference and ability to directly evaluate their probability density.
There are, however, other NFs architectures that have shown competitive performance in various generative tasks \cite{NeuralSplineFlows}.

Finally, our previous research has shown how NFs can be used as a complement to improve the performance of other models by providing a robust measure of confidence \cite{SimonSCIA, SimonIROS}.
It is possible that NFs could be applied in conjunction with for example DP or a transformer-based model to further improve performance.

\section*{ACKNOWLEDGMENT}

This research is funded by the Excellence Center at Linköping-Lund in Information Technology (ELLIIT), and the Wallenberg AI, Autonomous Systems and Software Program (WASP). We thank the computations enabled by the Berzelius resource provided by the Knut and Alice Wallenberg Foundation at the National Supercomputer Centre. We thank Hashim Ismail and Davide Tateo for their discussion and feedback, and Marcus Klang for continued support on robotic setup.

\bibliographystyle{IEEEtran}
\balance
\bibliography{IEEEabrv,main}

\end{document}